\documentclass{iaesarticle}

\usepackage{amsmath, amsfonts, amssymb, float, fancyhdr}
\usepackage[figuresright]{rotating}
\usepackage{authblk, graphicx, indentfirst, lastpage, lipsum}
\usepackage{pifont}

\makeatletter
\renewcommand{\@biblabel}[1]{[#1]\hfill}
\renewcommand\AB@authnote[1]{\textsuperscript{\normalfont\bfseries#1}}
\makeatother
\usepackage[font=normalsize]{subfig, caption}
\usepackage{epstopdf}
\usepackage[left=3cm, right=2.5cm, top=2.5cm, bottom=2.5cm, includehead, includefoot]{geometry}
\usepackage[justification=centering]{caption}
\usepackage{titlesec}
\usepackage{xcolor}
\titleformat{\section}
{\normalfont\normalsize\bfseries\uppercase}{\thesection}{1.7em}{}
\titleformat{\subsection}
{\normalfont\normalsize\bfseries}{\thesubsection}{.95em}{}
\titleformat{\subsubsection}
{\bfseries}{\thesubsubsection}{.2em}{}
\titlespacing*{\section}{0cm}{0.7cm}{0cm}
\usepackage{enumitem}
\usepackage[numbers,sort&compress]{natbib}
\usepackage{ragged2e}
\usepackage{hyperref,graphicx}
\usepackage{multirow}

\CopyrightLine[]{}{\textit{This is an open access article under the \textcolor{blue}{\underline{CC BY-SA}} license.}
\vspace{.5em}}

\author[1]{\bfseries Bousselham EL HADDAOUI}
\author[1]{\bfseries Raddouane CHIHEB}
\author[1]{\bfseries Rdouan FAIZI}
\author[1]{\bfseries Abdellatif EL AFIA}

\affil[1]{NATIONAL HIGHER SCHOOL FOR COMPUTER SCIENCE AND SYSTEMS ANALYSIS (ENSIAS), MOHAMMED V UNIVERSITY IN RABAT, MOROCCO}

\title{Sentiment Analysis in SemEval: A Review of Sentiment Identification Approaches}
\shorttitle{Sentiment Analysis in SemEval: A Review of Sentiment Identification Approaches (B. EL HADDAOUI)}

\begin{document}
\setcounter{page}{1}

\setlength{\parindent}{1.27cm}

\pagestyle{fancy}
\fancyhfoffset{0cm}

\journalname{International Journal of Electrical and Computer Engineering (IJECE)}
\journalshortname{Int J Elec \& Comp Eng}
\journalhomepage{http://ijece.iaescore.com}
\vol{99}
\no{1}
\months{Month}
\years{2099}
\issn{2088-8708}
\DOI{10.11591}
\pagefirst{1}
\pagelast{1x}
\IDpaper{paperID}

\maketitle

\hrule
\vspace{.1em}
\hrule
\vspace{.5em}
\noindent
\parbox[t][][s]{0.315\textwidth}{%
\textbf{Article Info}
\vspace{.5em}
\hrule
\vspace{.5em}
\begin{history}
\vspace{.5em}

Received Jun 8, 2022

Revised Jul 16, 2022

Accepted Aug 18, 2022

\vspace{.7em}
\end{history}
\vspace{.5em}
\hrule
\vspace{.5em}
\begin{keyword} 
\vspace{.5em}
Sentiment Analysis \sep Machine Learning \sep Deep Learning \sep Transformers \sep Social Media
\vspace{.5em}
\end{keyword}
\vspace{\fill}
}
\parbox{0.020\textwidth}{\hspace{0.5em}}
\parbox[t][][s]{0.65\textwidth}{%
\begin{abstract}
\vspace{.3em}
Social media platforms are becoming the foundations of social interactions including messaging and opinion expression. In this regard, Sentiment Analysis techniques focus on providing solutions to ensure the retrieval and analysis of generated data including sentiments, emotions, and discussed topics. International competitions such as the International Workshop on Semantic Evaluation (SemEval) have attracted many researchers and practitioners with a special research interest in building sentiment analysis systems. In our work, we study top-ranking systems for each SemEval edition during the 2013-2021 period, a total of 658 teams participated in these editions with increasing interest over years. We analyze the proposed systems marking the evolution of research trends with a focus on the main components of sentiment analysis systems including data acquisition, preprocessing, and classification. Our study shows an active use of preprocessing techniques, an evolution of features engineering and word representation from lexicon-based approaches to word embeddings, and the dominance of neural networks and transformers over the classification phase fostering the use of ready-to-use models. Moreover, we provide researchers with insights based on experimented systems which will allow rapid prototyping of new systems and help practitioners build for future SemEval editions.
\end{abstract}
}
\parbox[l]{\textwidth}{%
\rule{0.275\textwidth}{0.5pt} \hspace{0.5cm} \hrulefill
\\
\emph{\textbf{Corresponding Author:}}
\vspace{.5em}\\
Bousselham EL HADDAOUI\\
ENSIAS, MOHAMMED V UNIVERSITY IN RABAT, MOROCCO\\
Email: bousselham.haddaoui@um5s.net.ma
}
\vspace{.5em}
\hrule
\vspace{.1em}
\hrule

\section{Introduction}
\label{sec:introduction}

Technological growth has shaped, by different means, human interactions. Social media platforms, forums, and news websites are providing digital services allowing internet users to perform daily activities such as information sharing, messaging, and public discussions. With a 4.65B active social media users representing 58.7\% of the world population estimated at 7.89B  \cite{ref0}, industry stakeholders (i.e. e-commerce, politics, healthcare, etc.) show interest in the study of the large generated data from user interactions to extract insights and business value for decision-making, urging the need for the design of suitable techniques to process the available data with reasonable time and cost investment considerations.  In this regard, Sentiment Analysis, which is an interdisciplinary field that focuses on the extraction and analysis of information from diverse data sources (i.e., short and large text, images, and videos), along with Natural Language Processing (NLP), computational linguistics, and text mining techniques have provided the foundations to design automated and industry-grade systems to address these challenges \cite{ref1}.  Sentiment analysis focuses on various tasks including sentiment classification, aspect extraction, topic modeling, etc.  It has evolved from a simple classification task to a problem of automatic opinion detection and opinion-related attributes identification \cite{ref2} profiting from advancements in neural networks and other emerging research trends such as Pre-trained Language Models (PLM) \cite{ref2-1} and Neuro-Symbolic AI \cite{ref2-2}. Currently, sentiment analysis systems are widely used in the social media context and have many applications in other domains such as healthcare, social sciences, market research, political science \cite{ref3}, etc. 

Research studies, in this field, usually cover one or two aspects (i.e., word representations, features extraction, classification, etc) which result in knowledge sparsity. The need for automated systems motivated a research interest shifting toward the sentiment analysis pipeline’s design and implementation, thus providing state-of-the-art baselines. In this respect, international competitions such as the International Workshop on Semantic Evaluation (SemEval) are committed to raising awareness about sentiment analysis in the industry and academic communities. They have focused on the study of various social phenomena through their editions. SemEval organizes several sub-tasks covering trending sentiment analysis topics such as sentiment and emotion detection and classification, social phenomena such as hate speech, sarcasm, and offense, named entity recognition, etc. and ensures the evaluation of the submitted systems. Furthermore, the submitted systems are ready-to-use technical implementations of sentiment analysis systems that are based on recent research trends, thus providing a variety of system choices and fostering the improvement of previously submitted systems.

In our study, we aim to provide an in-depth analysis of submitted systems through the SemEval editions marking the evolution of used techniques on various sentiment analysis aspects, highlight the opportunities and limitations of proposed approaches, and present future research trends of sentiment analysis systems  implementations. In this respect, we review the five top-performing systems for each edition from 2013 to 2021 to provide insights regarding the evolution of used datasets, their preparation, and the annotation process. Moreover, we present the main used techniques and research innovations concerning various sentiment analysis systems aspects including data acquisition, data preparation, preprocessing techniques, features engineering, and classification approaches.

The rest of the paper is structured as follows. Section~\ref{sec:history_semeval} presents the history of the SemEval competition and a description of the challenges and research questions for each edition. Section~\ref{sec:proposed_systems} provides a review and analysis of the studied systems focusing on defined aspects such as datasets, preprocessing techniques, word representation and features engineering, and classification models.  A timeline tracking the evolution of used techniques for each aspect was also provided highlighting key contributions to the sentiment analysis field. In section~\ref{sec:findings}, key findings are presented and discussed. Conclusions are presented in section~\ref{sec:conclusion}.

\section{History of SemEval Tasks}
\label{sec:history_semeval}

The SemEval competitions focus on various NLP research topics including text similarity and question answering, time and space, word sense disambiguation and induction, learning semantic relations, sentiment analysis, etc. In our study, we considered the sentiment analysis special track and narrowed our coverage to the sentence classification task. The timeline of the covered SemEval editions, the subject of our study, is presented below:\\

\textbf{- SemEval-2013 Task 2}\\
In  \cite{ref4}, the first edition of the competition, the subtasks focused on the contextual polarity disambiguation and the message polarity classification. The key challenges for these tasks included the lack of suitable datasets for training and submissions assessment, the informal nature and the creative content retrieved from Twitter, and the overall sentiment conveyed in messages containing opposite sentiments. A total of 44 teams participated in the task.

\textbf{- SemEval-2014 Task 9}\\
In the second edition  \cite{ref5}, the organizers deepened the research questions for the previous subtasks by introducing two additional datasets covering formal content and sarcasm social content. The main challenges for these tasks included new content patterns such as the use of creative spelling and punctuation, slang, new words, and abbreviations. Sarcasm handling was the key addressed topic along with defining sentiment strength in tweets conveying opposite sentiments. A total of 46 teams participated in the task.

\textbf{- SemEval-2015 Task 10}\\
The 3rd edition of the competition  \cite{ref6} introduced new challenges related to sentiment analysis. In addition to the previous two subtasks, message sentiment and overall trend toward a topic were investigated along with the strength of association of specific words and positive sentiment. Challenges for these tasks included new patterns in social content such as emoticons, acronyms, and poor grammatical structure. A total of 40 teams participated in the task.

\textbf{- SemEval-2016 Task 4}\\
In  \cite{ref7}, the 4th edition of the competition, the sub-tasks introduced the ordinal classification instead of the usual binary classification for the previous massage polarity and sentiment toward topic tasks, and the quantification task which implies the study of the distribution of classes in unlabeled datasets. The key challenges for these tasks are the lack of training datasets and class imbalance, the difficulty of handling multi-class ordinal classifications, and the evaluation metrics for the quantification task. A total of 43 teams participated in the task.

\textbf{- SemEval-2017 Task 4}\\
The 5th edition  \cite{ref8} was a rerun of the previous competition, the organizers initiated the cross-language sentiment analysis with the introduction of the Arabic language for all the subtasks. In addition, the datasets provided user-related demographic information such as age, gender, location, etc. for usage as extra features. The main challenges for these tasks involved the lack of training data for the Arabic language, the high featured level of the language, and the abundant use of dialect forms and spelling variants. Besides, the cross-language nature of the ordinal classification and topic-related sentiment extraction. A total of 48 teams participated in the task.

\textbf{- SemEval-2018 Task 1}\\
In  \cite{ref9}, the 6th edition of the competition, the organizers focused on inferring the mental state of the user. In this respect, the subtasks covered the emotional intensity and the valence ordinal classification and regression, and emotion classification in a cross-language context. The limited training datasets and the cross-language were the main challenges in the preparation for this task. A total of 75 teams participated in the task marking an increased interest in the competition.

\textbf{- SemEval-2019 Task 9}\\
In the 7th edition  \cite{ref10}, the competition focused on offensive language identification and categorization. The proposed tasks included offensive language detection, categorization, and offense target identification. Challenges for these tasks included the process of building the evaluation dataset which should overcome the annotator’s bias and an extensive understanding of offensive language categorization. A total of 155 teams participated in the task showing a growing interest in the task.

\textbf{- SemEval-2020 Task 12}\\
In  \cite{ref11}, the 8th edition of the competition, the organizers introduced a multilingual aspect for offense identification and categorization. The provided dataset followed the best practices in abusive language collection  \cite{ref12} and covered five languages which are Arabic, Danish, English, Greek, and Turkish. Besides the challenges from the previous edition, the targeted offense presented many challenges including implicit and explicit offenses for individual and group targets that could be based on gender, religious beliefs or ethnicity, etc. A total of 145 teams participated in the task.

\textbf{- SemEval-2021 Task 7}\\
In the 9th edition  \cite{ref13}, the scope of the competition investigated both humor and offense through 4 subtasks which are humor detection, humor rating and controversy prediction, and offense detection. The main challenges in this competition included the ability to differentiate between humor and offense in tweets, and the perception of humor that can vary depending on age, gender, personality, etc  \cite{ref14}. Moreover, the levels of controversy in judgments between interceptors were also a challenge to address in these subtasks. A total of 62 teams participated in the task, the competition organized 11 different tasks.

SemEval competitions addressed many active sentiment analysis topics through their tasks, starting from the basic sentence and message polarity, message polarity toward a topic, to more advanced topics such as the study of figures of speech (i.e. sarcasm, offense, and humor), cross-language and ordinal classification. Moreover, notable limitations were encountered including the difficulty of preparing suitable datasets for training and evaluation purposes and the specificity of social media language. The proposed systems, in the previous competitions, provided ready-to-use solutions to answer the research questions marking a timeline of the evolution of tools and methodologies used in the sentiment analysis research area.

\section{Proposed systems}
\label{sec:proposed_systems}

\subsection{Datasets}

Since 2013, the competition has focused mainly on datasets collected from Twitter \cite{ref14-1}.  Alternative sources were considered including Reddit, News, and Kaggle to enrich the main dataset depending on the chosen topic. Data collection and annotation processes are required to prepare the competition datasets for training and evaluation purposes. The data collection phase is ensured by endpoints that are provided, by social media platforms, for content retrieval using open source software, otherwise scraping techniques or ready-to-use datasets are considered. Following this, organizers pull out content with no sentiment-bearing words to reduce class imbalance by keeping content with a score superior to 0.3 using SentiWordNet \cite{ref14-2} in the 2013-2018 editions and remove duplicates relying on Bag of Words (BOW) cosine similarity that exceeds 0.6 in 2018 edition.

The data annotation process is conducted after collection and cleaning depending on the task nature, various methodologies were considered through the editions including crowdsourcing platforms such as Amazon’s Mechanical Turk \cite{ref14-3},  CrowedFlower (rebranded to Appen \cite{ref14-4}), and Prolific \cite{ref14-5}. Furthermore, majority voting systems were used based on expert manual annotations in  \cite{ref10}. An initiative to use semi-supervised annotation systems based on ensemble techniques was introduced in  \cite{ref11}, outputs from selected voters including Pointwise Mutual Information (PMI)  \cite{ref15}, Fast Text  \cite{ref16}, Long Short-Term Memory (LSTM) networks  \cite{ref17}, and Bidirectional Encoder Representations from Transformers (BERT)  \cite{ref18}, are combined to assess the agreement score between voting models using Fleiss’ K Inter-Annotator Agreement (IAA)  \cite{ref19}.\\
\begin{figure}[H]
\centering
\includegraphics[width=10cm]{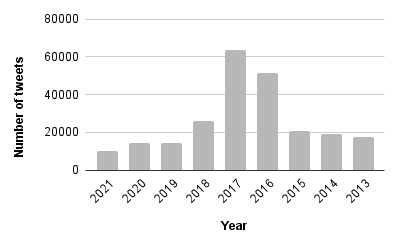}
\vspace{.7em}
\caption{The evolution of datasets size during the SemEval competition}
\label{Fig:datasets_evolution}
\end{figure}

The manual aspect of the preparation of the datasets makes it extremely slow and very expensive, thus the output of this process results usually in small to medium size datasets. The lack of suitable, class imbalance, and insufficient training datasets was the main restriction for the early editions considering the dependence of proposed systems on linguistic resources  \cite{ref20} and the limited datasets improvement which doesn’t impact the overall score during the evaluation process  \cite{ref5}.

The prepared datasets size depends on the nature of the competition topics,  and available datasets are reused and enriched by organizers for the reconducted or similar tasks as detailed in Table~\ref{Tab:datasets_stats}. Moreover, new datasets are prepared for specific tasks such as humor, offense, racism, etc,  and for which the sentiment distribution isn’t described since organizers used different classes (i.e., offensive/not offensive, humor/not humor,  emotions ordinal classification).  As for the datasets evolution, the organizers collected 17 401 tweets in 2013 to reach 63 677 tweets in 2017 as presented in Figure~\ref{Fig:datasets_evolution}, while new datasets were manually prepared for specific tasks from 2018 to 2021. The SOLID dataset (9 093 037 tweets) used in the 2020 edition \cite{ref11} which was prepared using a semi-supervised technique \cite{ref21} wasn’t considered in Figure~\ref{Fig:datasets_evolution} to appreciate the evolution of original datasets, manually collected, and annotated by experts.

\begin{table}[H]
\centering
\fontsize{8pt}{10pt}\selectfont
\caption{SemEval 2013-2021 Datasets Statistics}
\begin{tabular}{crlrrr}
\hline
 \multirow{2}{*}{\textbf{Edition}} &
  \multicolumn{1}{c}{\multirow{2}{*}{\textbf{Dataset size}}} &
  \multirow{2}{*}{\textbf{Data source}} &
  \multicolumn{3}{c}{\textbf{Sentiment distribution}} \\
 &
  \multicolumn{1}{c}{} &
   &
  \multicolumn{1}{l}{\textbf{Positive}} &
  \multicolumn{1}{l}{\textbf{Negative}} &
  \multicolumn{1}{l}{\textbf{Neutral}} \\
\hline
\multirow{2}{*}{2021} &
  8,000 &
  Twitter &
  \multicolumn{3}{l}{\multirow{5}{*}{ }} \\
                      & 2,000     & Kaggle       & \multicolumn{3}{l}{}     \\
2020                  & 9,107,137 & Twitter      & \multicolumn{3}{l}{}     \\
2019                  & 14,100    & Twitter      & \multicolumn{3}{l}{}     \\
2018                  & 26,184    & Twitter      & \multicolumn{3}{l}{}     \\
2017                  & 62,617    & Twitter      & 22,277 & 11,812 & 28,528 \\
\multirow{3}{*}{2016} & 50,158    & Twitter      & 19,855 & 9,004  & 21,299 \\
                      & 2,093     & SMS          & 492    & 394    & 1,207  \\
                      & 1,142     & Live journal & 427    & 304    & 411    \\
\multirow{3}{*}{2015} & 19,526    & Twitter      & 7,864  & 3,014  & 8,648  \\
                      & 2,093     & SMS          & 492    & 394    & 1,207  \\
                      & 1,142     & Live journal & 427    & 304    & 411    \\
\multirow{3}{*}{2014} & 17,134    & Twitter      & 6,824  & 2,649  & 7,661  \\
                      & 2,093     & SMS          & 492    & 394    & 1,207  \\
                      & 1,142     & Live journal & 427    & 304    & 411    \\
\multirow{2}{*}{2013} & 15,196    & Twitter      & 5,810  & 2,407  & 6,979  \\
                      & 2,094     & SMS          & 492    & 394    & 1,208    \\                              
\hline
\end{tabular}
\label{Tab:datasets_stats}
\end{table}

To overcome the limitations of size and quality in the competitions datasets, participants further enriched the training data by introducing public datasets from known repositories such as Kaggle \cite{fn7},  UCI \cite{fn8},  Github \cite{fn9}, etc., and collected social media content from social platforms using topic related word seeds  \cite{ref13} or emoticons  \cite{ref22} as keywords for queries.

\subsection{Data Preprocessing}

Social media content, the primary focus of SemEval competitions, gained big data attributes such as volume, variety, and velocity over time. Furthermore, the rich nature and shape of social messages which hold information about individuals and their interactions  \cite{ref4}, and convey user sentiment  \cite{ref23} can be altered by different means considering the creative and informal social content. Misspellings, poor grammatical structure, hashtags, punctuation, new words, emoticons, acronyms, and slang  \cite{ref6} are language phenomenons qualified as noise that causes extreme lexical sparsity  \cite{ref24}. To handle these language issues, preprocessing tasks had become an essential component of every sentiment analysis system. In addition, other complex methods such as word context disambiguation, and out-of-vocabulary (OOV) handling can be applied at this level to improve content quality and system performance  \cite{ref25}. 

The need for automation for sentiment analysis systems  \cite{ref20} gives grounds for advanced use of preprocessing techniques and raises research interest in their usability and effectiveness in the context of social media. The need to clean noisy data  \cite{ref4} without affecting text meaning is the main concern of preprocessing, especially for social media content whose features are linguistically driven and require specific text processing  \cite{ref26}. Moreover, text normalization techniques including hashtags segmentation, emojis conversion, and spell correction  \cite{ref11} have shown their effectiveness in dealing with large vocabularies  \cite{ref27}, negation handling  \cite{ref28}, and OOVs  \cite{ref29}. The effectiveness of some preprocessing techniques such as punctuation removal is still controversial since punctuation may not affect the general meaning, while in other cases (i.e. multiple exclamations), might be useful as additional features  \cite{ref30}. \\

\begin{figure}[H]
\centering
\includegraphics[width=10cm]{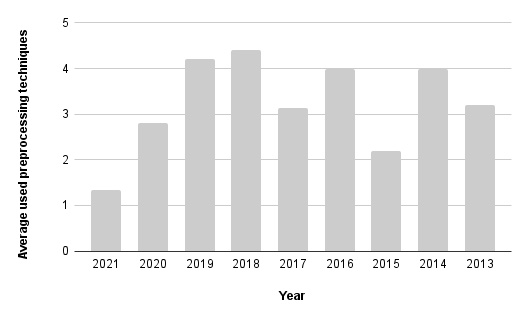}
\vspace{.7em}
\caption{The Average used preprocessing techniques by edition}
\label{Fig:average_preprocessing}
\end{figure}

In our study scope, top-performing participants combined at least one to four preprocessing techniques as shown in Figure~\ref{Fig:average_preprocessing}, the appeal to preprocessing techniques is usually motivated by a will to improve state-of-the-art models. Whereas a decline in the use of preprocessing refers to the introduction of new models or methodology. There was a similarity in used processing techniques between the 2013 and 2014 editions of the competition  \cite{ref5}, while no new trends were observed in the 2016 edition  \cite{ref7}. A particular use of ready-to-use text preprocessors such as the Ark Tokenizer  \cite{ref31} and the NLTK TweetTokenizer \cite{fn10}, Stanford Core NLP  \cite{ref32}, and Keras \cite{fn11} by participants was noted in the 2019 edition with a focus on the combination of preprocessing techniques in the 2021 edition. The review of used techniques through the considered systems allowed us to perform a summarization by usage frequency, below is a list of used
preprocessing techniques: 
\begin{itemize}
\item \textbf{High usage frequency:} Tokenization,  Lowercasing, URL, User mentions, and Special characters (Unicode, XML, etc.) removal. 
\item \textbf{Average usage frequency:} Parts of Speech (POS), Convert emojis to words, Duplicates, Punctuation, and Hashtag removal.
\item \textbf{Low usage frequency:} Lemmatization, Negation handling, Spelling correction, Stemming, Truncate tweet, Remove stop words and extra white space.
\end{itemize}

\begin{figure}[H]
\centering
\includegraphics[width=10cm]{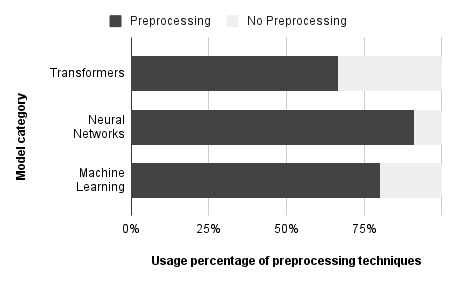}
\vspace{.7em}
\caption{Usage of preprocessing techniques per sentiment analysis model category}
\label{Fig:average_preprocessing_model}
\end{figure}

The usage of preprocessing techniques depends on the analysis approach, some techniques remove useless content while others improve classifier performance  \cite{ref33}. The studied systems used various approaches such as Machine Learning classifiers, Neural Networks, and Transformers. Figure~\ref{Fig:average_preprocessing_model} shows usage ratios of preprocessing techniques by classification approach, transformer-based approaches rely less on preprocessing techniques compared to other approaches which consider it a key component of every sentiment analysis pipeline.

\subsection{Data representation and features engineering}

Sentiment can be expressed in different forms on social media, and in most cases in a discrete  \cite{ref34} and an implicit way  \cite{ref35} which makes the identification and extraction tasks difficult to perform. The complexity of human sentiment can vary from basic emotions, described in the Basic Emotion Model, such as joy, sadness, and fear that can be physiologically and cognitively expressed  \cite{ref36}  \cite{ref37}, to more complex ones studied in the Valence-Arousal-Dominance (VAD) Model  \cite{ref38} which categorize emotions into a dimensional grouping. Valence (positiveness-negativeness), arousal (active-passive), and dominance (dominant, submissive) are dimensions of the emotion space where sentiments can be represented. Furthermore, a sentiment representation was proposed in  \cite{ref35} into a machine-understandable form consisting of an entity E(j), the aspect of the entity A(j,k), the sentiment S(j,k,i,l), the sentiment holder H(i), and the time T(l). In addition, the geographical location was incorporated to enrich the representation in  \cite{ref39}.

State-of-the-art systems depend heavily on linguistic resources, extensive features engineering, and tuning  \cite{ref20}. Moreover, sentiment categorization relies on feature choice which remains a primary challenge for the sentiment analysis systems  \cite{ref26}, and feature selection strategies (i.e. linguistic, lexical, or mixed strategy). In  \cite{ref4}, most systems proposed a lexicon-based strategy while pointing to a lack of sentiment lexicons  \cite{ref40}. Available resources, as presented in  \cite{ref29}  \cite{ref40}  \cite{ref23}  \cite{ref5}  \cite{ref41}  \cite{ref42}  \cite{ref9}, included formal lexicons such as General Inquirer  \cite{ref43}, MPQA Subjectivity Lexicon  \cite{ref44}, SentiWordNet  \cite{ref45}, and informal ones such as AFINN-111  \cite{ref46}, Bing Liu’s Opinion Lexicon  \cite{ref47}, NRC Hashtag Sentiment Lexicon  \cite{ref48}, Sentiment140 Lexicon  \cite{ref49}. Other lexicon datasets were used by participants including DeRose \cite{fn12}, Urban Dictionary slang dictionary, and large-vocabulary distributional Semantic models (DSM) constructed from Wacky web-crawled large corpora  \cite{ref50} and the Google Web 1T 5-Grams database \cite{fn13}. In addition, research efforts focused on improving proposed systems with manually created domain-dependent and independent taxonomies from open data sources such as forums, news, etc. which require in some cases domain knowledge  \cite{ref51} or based on sentiment-bearing word seeds including hashtags, emoticons, keywords, etc. form social media platforms  \cite{ref40}.

Although lexicon-based approaches contribute effectively to determining the overall sentiment, additional feature engineering may be required to further improve model performance. In this respect, specific features of social context are considered including Twitter profile’s demographic information such as user age, gender, location, followers count, etc.)  \cite{ref8}, other approaches are considered such as word sense disambiguation  \cite{ref29}, negation handling  \cite{ref28}, and emojis conversion  \cite{ref52}. The focus on features engineering is motivated by the contribution of rich features set on the classification models  \cite{ref23},  and feature weighting may reflect the influence of each feature on the overall sentiment  \cite{ref24}.

In  \cite{ref9}, three main feature engineering techniques were highlighted including lexicon features, word n-grams, and word embeddings. Although important features are extracted from lexicons  \cite{ref6}, lexicon-based approaches present many limitations resulting essentially from manual annotation effort, domain-dependent features crafted from sentence words, emotions, slang, hashtags, etc.  \cite{ref30} which motivated word-based features engineering. In word-based approaches, a variety of techniques were used by participants in  \cite{ref4} such as word-based (i.e. stems, words, clustering, n-grams), word-shape (i.e. capitalization, punctuation), syntactic (i.e. dependency relations, part of speech tags (POS)), and Twitter-specific features (i.e. emoticons, repeated letters, hashtags, URLs, abbreviations, and slang). Moreover, surface form features can be also considered including the number of elongated words, the number of hashtags  \cite{ref40}, expanded words from a predefined word list  \cite{ref51}, word shape, interjection (words that express a sentiment such as lol, hurrah)  \cite{ref23}, the number of tokens in a sentence  \cite{ref30}, negation and semantic features  \cite{ref41}, all-caps and punctuation  \cite{ref53}. Another explored approach in this context is Bag of Words (BOW)  \cite{ref54}, various sets of bags of words including unigrams, bigrams, and extended unigrams models were explored in  \cite{ref30}. The approach is one of the most important representation methods  \cite{ref55}, and it becomes less effective in short texts and leads to increased data sparseness. In  \cite{ref27}, additional techniques are provided to complement BOW with a denser representation including weighting relevant words in a BOW (BM25)  \cite{ref56}, mapping words to clusters using Brown Clusters  \cite{ref57}, or assigning weight to terms on each identified class using Concise Semantic Analysis  \cite{ref58}. The performance yielded by lexicons and word-based approaches is coupled with a difficulty in relying on manual features, thus the need for a new approach  \cite{ref22}.

Participants in recent SemEval editions focused on the latest research trends through the use of unsupervised learning of word embeddings  \cite{ref20}. Word2vec  \cite{ref59} and Global Vectors (GloVe)  \cite{ref60} are two frequently used unsupervised word embedding techniques that provide general-purpose and multidimensional word embeddings and provide fine-tuning mechanisms for domain-dependent data representation  \cite{ref22}. Furthermore, improvements to the previous techniques are used such as FastText  \cite{ref61} which enhances Word2vec’s ability to train on small datasets and generalize to unknown words, sentiment embeddings fusion, and sentiment-specific word embeddings that consider word sentiment in addition to syntactic and semantic contexts  \cite{ref62}. Moreover, character embeddings were explored to overcome the limitations of word embeddings in handling OOVs  \cite{ref25}, while traditional approaches usually ignore OOVs (i.e. set to default or neutral words) ignoring the eventuality of being sentiment-bearing words. In the 2020 edition, systems relied more on contextualized word representations provided by transformers such as BERT  \cite{ref18}, ELMo  \cite{ref63}, RoBERTa  \cite{ref64}, and Multilingual BERT (mBERT), due to the significant improvements brought to various NLP tasks  \cite{ref65}.

\subsection{Classification models}

SA systems give special consideration to the classification task as it represents the core component of the majority of proposed systems. Classification approaches fall, generally, under three categories including supervised, semi-supervised, and unsupervised techniques. The first editions of SemEval noted a trend focusing on supervised learning  \cite{ref6}, models such as Support Vector Machines (SVM), Naive Bayes (NB), Maximum Entropy, Rule-based classifiers, and ensemble techniques  \cite{ref4} were the most used. Whereas, recently top-performing systems relied more on neural networks, language models, and features derived from existing emotion and sentiment lexicons  \cite{ref9}. Organizers paid special attention to constrained (relying only on provided datasets) and unconstrained systems (using additional resources) which allowed the use of transductive (the same task for different domains) and inductive (the same domain for different tasks) learning strategies. Four used model classes were identified during our study, categories include Machine Learning (ML), Deep Learning (DL), Transformers, and Ensemble models.

Initial approaches for classification were rule-based  \cite{ref4}. Rules matching may cover words and sequences, and are either handwritten, identified using statistical methods, or available software such as Synesketch  \cite{ref66}. Although they have high precision, rule-based systems require skilled linguists to define rules and lack generalization and scalability which are addressed and improved with machine learning algorithms  \cite{ref67}. Machine learning models provided state-of-the-art results during the 2013-2016 editions, SVM, Maximum Entropy, Conditional Random Fields (CRFs), Linear Regression, and Logistic Regression were the most commonly used models  \cite{ref6}. Models were enhanced using cross-validation on training datasets  \cite{ref40}, using loss functions such as Hinge Loss Function to improve accuracy, and regularization such as L1, L2, or Elastic Net Regularization (combination of L1 and L2)  \cite{ref26} to avoid overfitting. ML-based approaches have known limitations related to poor transfer learning and limited ability to learn complex patterns which motivated the use of DL approaches in a quest to further improve systems performance. 

Since SemEval 2015, top-performing systems have used mainly DL models built using deep neural networks and word embeddings  \cite{ref6}. The limitation of transfer learning in ML models can be overcome by initializing neural networks with an embeddings layer pre-trained on available general-purpose or domain-specific corpus (i.e. GloVe 25d, 50d, 100d, and 300d, Google new corpus with 300d vectors, etc.)  \cite{ref68}, networks can be also initialized and refined using distant learning as in  \cite{ref49} where a convolutional neural network (CNN) inspired from  \cite{ref69} was used for this purpose  \cite{ref70}. This usage trend was manifested in SemEval 2019  \cite{ref10} with more than 70\% of systems using various DL models including LSTM, Bidirectional LSTM (BiLSTM), Recurrent Neural Networks (RNN), CNN, and Gated Recurrent Unit (GRU) Neural Networks. Although DL systems proved their high performance, representation learning techniques (based on DL neural networks) can benefit from manually engineered features  \cite{ref10}. Furthermore, additional optimization techniques were considered such as adding small perturbations on input samples using adversarial examples  \cite{ref71} which proved to improve models loss  \cite{ref72} along with advanced techniques that may be used including Grid Search, Random Search, and Genetic Algorithm to learn neural networks hyperparameters  \cite{ref73}. 

Driving a paradigm shift in the use of DL approaches, Pre-trained Language Models (PLMs) or Transformers are the new evolution of the neural network models. BERT  \cite{ref18} has shown significant performance and proved resistant to overfitting compared to other models  \cite{ref74}, and it helps better understanding of sentence meaning and generates expressive word-level representations due to the inherent data noise in social media content  \cite{ref75}. Furthermore, PLM such as BERT and Universal Language Model Fine-tuning (ULMFiT) \cite{ref750} provided better results in multilingual classification  \cite{ref13} compared to systems based on linear models and optimized using evolutionary algorithms. New PLMs were introduced that outperform existing models to optimize existing models in various directions such as providing lite versions that reduce parameters, increase models speed, and reduce memory consumption (i.e. ALBERT \cite{ref751} for BERT). Other alternatives focused on training strategies such as XLNet \cite{ref752} which introduced an automatic regressive pre-training method  \cite{ref76}. This notable advancement in performance using transformers comes with many limitations including a negative impact of class imbalance  \cite{ref77}, and the required intensive computation using large PLM models  \cite{ref78}.

Ensemble learning was also considered the SemEval systems combining top-performing ML and DL models  \cite{ref79} provided a hybrid approach taking the benefits of each model category. The weighting scheme varies depending on the adopted strategy, soft voting  \cite{ref68} and hard majority voting  \cite{ref76} are known options that can be improved using algorithms (i.e. Limited-memory BFGS) to optimize weights attributed to each model.\\

\begin{figure}[H]
\centering
\includegraphics[width=10cm]{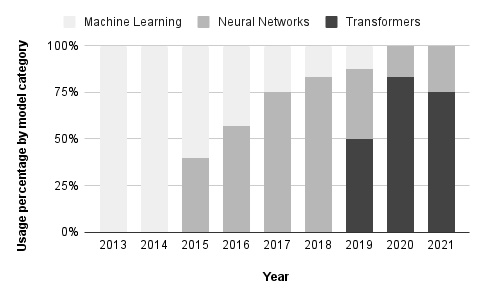}
\vspace{.7em}
\caption{Classification model usage evolution}
\label{Fig:classification_models}
\end{figure}

During the 2013-2021 SemEval editions, the evolution of models used in sentiment analysis-related tasks was observed. Contributions evolved using ML and DL models to recent Transformers following research trends to improve the performance and accuracy of proposed systems. Figure~\ref{Fig:classification_models} shows the evolution of usage per mode category backed by insights and taking into consideration the competition timeline and various constraints. The dominance of ML systems was challenged after the 2015 edition by the introduction of DL systems, which after the 2019 edition were less used compared to new Transformer-based systems. Our review allowed us to summarize the proposed systems by usage frequency through the 2013-2021 SemEval editions, below is the result of our categorization:
\begin{itemize}
\item \textbf{High usage frequency:} LSTM, CNN
\item \textbf{Average usage frequency:} BERT, SVM
\item \textbf{Low usage frequency:} ALBERT, RoBERTa, XLM-ROBERTa, ERNIE, RNN, GRU, XGBoost, Random Forest, Logistic regression, Naive Bayes, MLP, LibLinear, SGD, Rule-based
\end{itemize}

\subsection{Evaluation metrics}
The evaluation measures are determining factors in assessing the quality of sentiment analysis models. Model type, data variety, and size advise the evaluation metrics to be used  \cite{ref80}. In the SemEval competitions the F1-score was the most used evaluation metric (in 7 out of 9 studied competitions), other metrics were used including the average recall, the accuracy, and the Pearson Correlation Coefficient. The F1-score use was motivated by the strong imbalance between instances of different classes  \cite{ref10}, and where other measurements such as the accuracy fail. Complex metrics are used for model evaluation and analysis which provides a deep understanding of models behavior including the Confusion Matrix and the Receiver Operating Characteristic (ROC) curve.

\section{Findings and discussion}
\label{sec:findings}

SemEval remains an important sentiment analysis competition with new challenges and advanced complexity brought in every edition. On the macro-level, top-ranked systems tend to use universally effective approaches that deliver acceptable results  \cite{ref7}, researchers' efforts are then focused on fine-tuning models, features engineering, and datasets. Although the lack of training resources, and constrained systems relying only on the provided resources tend to perform better than unconstrained ones  \cite{ref5}, the reasons behind this lay in the additional datasets quality, the data labeling process, and the choice of the features. Moreover, the multilingual aspect of sentiment analysis was also explored noting the lack of models generalization due to the structural differences between languages (i.e. Arabic has abundant use of dialect forms and spelling variants) and the limited improvement using existing translation services  \cite{ref81}, which affect models performance and motivate the use of language-specific systems  \cite{ref8}.

On the micro-level, there are two approaches for features engineering in sentiment analysis systems: a lexicon-based approach that relies on lexicons to drive the polarity of text, and a machine learning approach that learns a classifying model from annotated text  \cite{ref41}. In the lexicon-based approaches, systems are constrained by the limited lexicon size and require human expertise; features engineered using this approach significantly impact the classifiers  \cite{ref24}. Although a richer feature set improves the classification task  \cite{ref23}, the overuse of features may lead to model overfitting  \cite{ref24}. Features based on lexicons have shown to be effective, furthermore, word sense features and word clusters are implemented to improve NLP systems  \cite{ref29}. In  \cite{ref26}, word clusters proved to be the most important to their model compared to n-grams and POS which didn’t add much improvement. Besides, the experiments in  \cite{ref27} show that Brown clusters yield the most considerable impact when compared to syntactic features and word unigrams, the laters leading to model performance degradation. In special cases, syntactic features such as punctuation (i.e. multiple exclamations)  \cite{ref30}, and negation patterns can be sentiment-bearing and affect the sentiment perception. Furthermore, Twitter-specific features may encode implicit sentiment and bring significant performance improvement  \cite{ref40} in the context of social media sentiment analysis which contains a lot of implicit sentiment.

In the ML-based approaches, word embeddings were demonstrated to be suitable as features compared to lexicon-based ones  \cite{ref53}, and scale well to complex language patterns  \cite{ref77}. Word embeddings, including general-purpose Word2vec or GloVe, are the most used, whereas few systems propose custom embeddings fine-tuned on provided datasets  \cite{ref7}. Moreover, the fine-tuning operation encodes additional semantic information and enriches the word representations. However, shows limited ability with domains containing high types and slang such as social media  \cite{ref25}. The embedding size affects the quality of results, large embeddings provide consistent results  \cite{ref20} while higher dimensions may not improve the performance  \cite{ref82}. Another limitation of word embeddings resides in handling out-of-vocabulary (OOV) words which can be partially overcome using character-level embeddings information  \cite{ref83}. Furthermore, starting from certain available training data, the choice of training models is more critical than the choice of features and word representations  \cite{ref26}.

Another important aspect of the studied systems is the classification models whose performance has improved over the editions, this can be explained by the advancement of learning methods and the amount of provided training data  \cite{ref7}. Most used approaches for sentiment detection involve methods from machine learning, computational linguistics, and statistics  \cite{ref79}, with a three-scale sentiment classification (positive, neutral, and negative)  \cite{ref26} or two points classification (i.e. positive/negative, subjective/objective, etc.). Specific classification categories such as subjectivity and objectivity can be more difficult to perform than the default positive and negative. Furthermore, classification from one scale can be the input features of a classification model for another scale (i.e. content with high positive/negative confidence is less likely to be objective, thus positive/negative output can be used to determine the objectivity/subjectivity)  \cite{ref42}.

Classification models differ in complexity, the most straightforward is rule-based relying on expert pattern detection and rules refinement. Besides, they can measure up to other approaches in performance and provide the flexibility of context-centric approaches such as sentiment toward a topic, the author’s feeling, or the general mood  \cite{ref51}. Linear models namely SVM and LR dominated the first editions and performed well compared to other models such as NB, RF, and KNN  \cite{ref26}; nevertheless, RF delivered a state-of-the-art performance  \cite{ref84}. The performance of those models can be further optimized using regularization techniques, this was observed in  \cite{ref79} where L2-regularized LR reimplementation has outperformed the original model. Emerging trends in  \cite{ref25} have manifested in the use of DL techniques, especially CNNs and RNNs, and the use of word embeddings such as GloVe and Word2vec. Moreover, the impact of word embeddings usage on the model’s performance was apparent as the initialization of networks with random parameters provides mediocre results while providing good initialization parameters (i.e. pre-trained word embeddings) boost the model’s performance  \cite{ref70}.  The massive use of LSTM and CNN implementations using Theano \cite{fn14} and Keras  \cite{ref7}, since their combination with word embeddings provided better performance and proved to be well suited for sentence classification problems. Furthermore, strategies of early and late fusion of word embeddings were explored noting limitations including a lack of features correlation modeling in the late fusion and an intensive required training for the early fusion  \cite{ref53}. The architectures of the proposed models were very basic, simply stacked layers of LSTM, BiLSTM, or CNN enhanced with attention, noise, dropout, and regularization techniques. Furthermore, the fine-tuning process focuses usually on selected hyperparameters such as the layers dimension, the number of layers, the learning rate, the epochs, etc.

In line with current research directions, recent systems use mainly PLMs such as BERT and its variants. PLMs brought noticeable advancement in performance compared to the DL model; a 10\% improvement in F1 score with BERT compared to LSTM was noted in  \cite{ref85}. Furthermore, PLM variants such as ALBERT (light BERT version) can yield comparable results to BERT  \cite{ref86} allowing a benefit of time and computational resources optimization, this can be enhanced using domain adaptation which can improve classification performance  \cite{ref87}. Another studied aspect is ensemble techniques which, depending on the models (voters), may lead to better results compared to individual systems  \cite{ref88}. Furthermore, ensemble techniques based on deep learning approaches profiting from the use of word embeddings and neural networks showed promising results  \cite{ref84}. Moreover, the use of stacking, as an ensemble strategy, demonstrated the robustness of systems  \cite{ref34} improving their accuracy  \cite{ref89}. Votes are decided using an average confidence model score, a straightforward scheme, or other strategies such as soft or hard voting  \cite{ref68}. The benefits of using such techniques come with exceptions where it shows performance degradation; an individual voter may outperform other group models which results in overall performance loss, voters should perform at a similar level (compatibility with the nature of the classification task) to ensure potential gains in performance considering the complexity and the intensive computation that come with a high cost of deployment of such models.

The evolution of used techniques in the proposed systems follows a logical path of research trends, and the transition from ML and DL models to current PLMs contributed to the performance improvement and the maturity of sentiment analysis-based systems to meet industry-grade requirements. While the appeal to transformers and neural networks was motivated by the considerable performance gain and the research context, the proposed systems provide less space for innovation and research efforts delegating this to the provided models and focusing on ready-to-use components. Moreover, the scope of intervention remains very limited for researchers to fine-tune pre-trained word embeddings or models, and basic architecture design which usually leads to minor gains in performance and notable complexity and computational cost. Another point to discuss is thematic sentiment analysis which explored, through SemEval competitions, the identification of some language phenomena such as sarcasm, offense, and humor. In this respect, specialized systems were built to benefit from domain fine-tuning and advanced features engineering for specific domains and languages, thus outperforming general-purpose models. The lack of training data and the difficulty of features engineering remains the main issues related to domain-specific models, thus encouraging the use of general-purpose models for rapid prototyping.

The proposed systems, in most cases, agreed on a similar tasks pipeline. Data cleaning and processing, feature engineering and representation, classification, and visualization, are the principal tasks performed by most participants. Besides, the focus was geared toward particular sub-tasks (i.e preprocessing, or classification) rather than the whole pipeline. The optimization process seems difficult to assert since the components are interdependent, poor word representation or the use of random preprocessing techniques may lead to overall performance degradation. Furthermore, raise concerns about the approaches that sentiment analysis systems should consider in the design phase, and urge the need for a common framework for sentiment analysis systems.

\section{Conclusion}
\label{sec:conclusion}

In this paper, we study top-ranked systems in the SemEval competition during the 2013-2021 period. The objectives are to provide a timeline tracking the evolution of used sentiment analysis techniques including data representation, preprocessing, and classification. We focused on the five top-performing systems in each edition, various aspects of these systems were analyzed including data preparation, preprocessing techniques, and classification models. Moreover, we provide a diagnosis and summarization of the challenges, key contributions, and limitations of each system to provide an overview of the sentiment analysis research field.

Notable progress was observed in the data preparation phase, organizers capitalized on the provided datasets from each edition and improved the preparation methodology from relying on manual labeling and expert work, the use of crowdsourcing platforms, to the use of techniques such as data augmentation and distant supervision which enrich datasets to overcome the lack of training datasets limitation noted in the first editions. Moreover, we pointed to the active use of preprocessing techniques in most studied systems with less frequency in transformer-based approaches, due to the specific nature of social media content and the gains in performance and computational complexity. Word embeddings such as GloVe and Word2vec remain the most used techniques for word representations and features encoding, state-of-the-art systems were provided by coupling word embeddings with neural networks. Furthermore, approaches relying on pre-trained language models provided comparable results with those systems and paved the way for their dominance over neural networks and traditional machine learning algorithms.

We believe our work will help researchers access a variety of experimented approaches for sentiment analysis systems in the context of social media, and allow future participants to provide more innovative solutions. In our future work, we intend to cover other submissions considering aspects such as innovation, and research originality to complement our top-performing filter. Moreover, thematic reviews for features engineering, preprocessing techniques, or classification approaches can be envisaged to provide an in-depth analysis of each sentiment analysis task.



\bibliographystyle{IEEEtran}


\section*{BIOGRAPHIES OF AUTHORS} 

\vspace{-.7em} 
\small
\begin{biography}[{\includegraphics[width=2.5cm,height=4cm,clip,keepaspectratio]{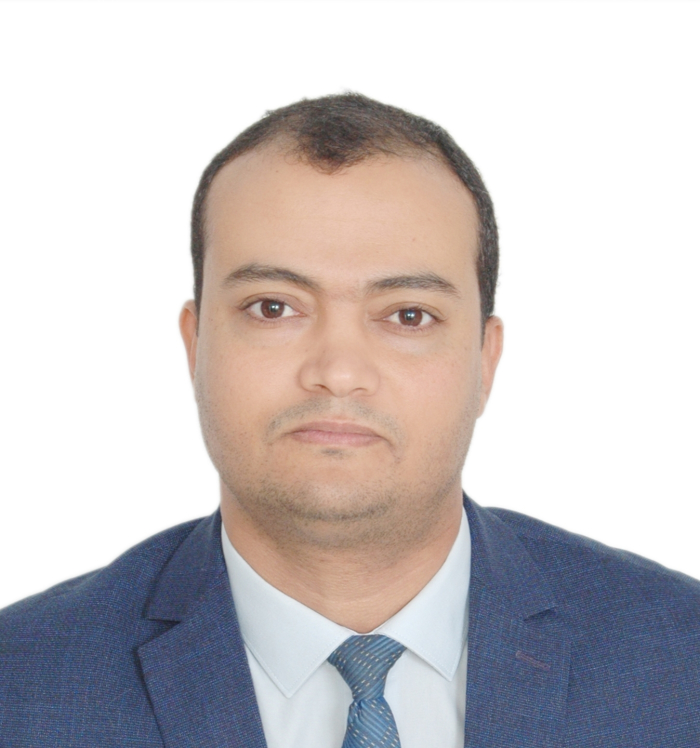}}]
\textbf{Bousselham EL HADDAOUI} 
  \href{https://orcid.org/0000-0002-9853-8794}{\includegraphics[width=0.02\textwidth]{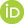}} 
\href{https://scholar.google.com/citations?user=Xi67OuMAAAAJ}{\includegraphics[width=0.02\textwidth]{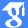}}
\href{https://www.scopus.com/authid/detail.uri?authorId=57203904879}{\includegraphics[width=0.02\textwidth]{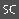}}
 is a Phd student specialized in Opinion Mining and Sentiment Analysis, Independent IT Advisor in digital communication and event management software. He has an engineering degree in software engineering from ENSIAS, Mohamed V University in Rabat Morocco, 2012.  He can be contacted at email: bousselham.haddaoui@um5s.net.ma.
\end{biography}

\vspace{-.7em} 
\small
\begin{biography}[{\includegraphics[width=2.5cm,height=4cm,clip,keepaspectratio]{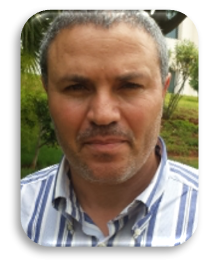}}]
\textbf{Raddouane CHIHEB} 
 \href{https://orcid.org/0000-0002-5297-2771}{\includegraphics[width=0.02\textwidth]{orcid.png}} 
\href{https://scholar.google.com/citations?user=2ZLdgNUAAAAJ}{\includegraphics[width=0.02\textwidth]{gscholar.png}}
\href{https://www.scopus.com/authid/detail.uri?authorId=6506589568}{\includegraphics[width=0.02\textwidth]{scopus.png}} 
 is a professor of Higher Education, Head of the Computer Science and Decision Support Department, PhD in Applied Mathematics from Jean Monnet University of Saint-Étienne, 1998. Specialized Master in Computer Science from Insa in Lyon, France, 2001, and the President of the Moroccan Association for Value Analysis.  He can be contacted at email: r.chiheb@um5s.net.ma.
\end{biography}

\vspace{-.7em} 
\small
\begin{biography}[{\includegraphics[width=2.5cm,height=4cm,clip,keepaspectratio]{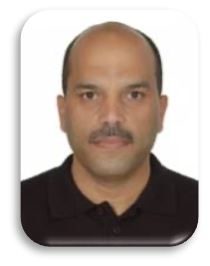}}]
\textbf{Rdouan FAIZI} 
  \href{https://orcid.org/0000-0003-4156-8113}{\includegraphics[width=0.02\textwidth]{orcid.png}} 
\href{https://scholar.google.com/citations?user=9HB4ZL8AAAAJ}{\includegraphics[width=0.02\textwidth]{gscholar.png}}
\href{https://www.scopus.com/authid/detail.uri?authorId=36989365300}{\includegraphics[width=0.02\textwidth]{scopus.png}} 
  is a Full Professor at National School of Computer Science and Systems Analysis (ENSIAS). He obtained his PhD in English Language and Literature from Mohammed V Agdal University, 2002. Research areas of interest are Linguistique, E-learning, Business English Skills, TOEIC Preparation, Scientific Communication. He can be contacted at email: r.faizi@um5s.net.ma.
\end{biography}

\vspace{-.7em} 
\small
\begin{biography}[{\includegraphics[width=2.5cm,height=4cm,clip,keepaspectratio]{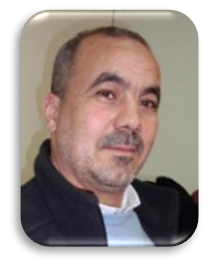}}]
\textbf{Abdellatif EL AFIA} 
  \href{https://orcid.org/00000-0003-1921-4431}{\includegraphics[width=0.02\textwidth]{orcid.png}} 
\href{https://scholar.google.com/citations?user=CLgOVHUAAAAJ}{\includegraphics[width=0.02\textwidth]{gscholar.png}}
\href{https://www.scopus.com/authid/detail.uri?authorId=56377420600}{\includegraphics[width=0.02\textwidth]{scopus.png}} 
   is a Full Professor at National School of Computer Science and Systems Analysis, He received his M.Sc.Degrees in Applied Mathematics from University of Sherbrook. He obtained his Phd in 1999 in Operation Research from the University of Sherbrook, Canada. Research areas of interest are Mathematical Programming (Stochastic and deterministic), Metaheuristics, Recommendation Systems and Machine Learning. He is the coordinator of the Artificial Intelligence Engineering branch (2IA) at ENSIAS.  He can be contacted at email: a.elafia@um5s.net.ma.
\end{biography}

\end{document}